# A WEIGHTED COMBINATION SIMILARITY MEASURE FOR MOBILITY PATTERNS IN WIRELESS NETWORKS

Thuy Van T. Duong[1], Dinh Que Tran[2] and Cong Hung Tran[3]

[1]Department of Information Technology, University of Ton Duc Thang,
Ho Chi Minh City, Vietnam
`vanduongthuy@yahoo.com`

[2]Faculty of Information Technology, Post and Telecommunication Institute of Technology, Km10, NguyenTrai, Hadong, Hanoi, Vietnam
`tdque@yahoo.com`

[3]Faculty of Information Technology, Post and Telecommunication Institute of Technology, Ho Chi Minh City, Vietnam
`conghung@ptithcm.edu.vn`

## ABSTRACT

*The similarity between trajectory patterns in clustering has played an important role in discovering movement behaviour of different groups of mobile objects. Several approaches have been proposed to measure the similarity between sequences in trajectory data. Most of these measures are based on Euclidean space or on spatial network and some of them have been concerned with temporal aspect or ordering types. However, they are not appropriate to characteristics of spatiotemporal mobility patterns in wireless networks. In this paper, we propose a new similarity measure for mobility patterns in cellular space of wireless network. The framework for constructing our measure is composed of two phases as follows. First, we present formal definitions to capture mathematically two spatial and temporal similarity measures for mobility patterns. And then, we define the total similarity measure by means of a weighted combination of these similarities. The truth of the partial and total similarity measures are proved in mathematics. Furthermore, instead of the time interval or ordering, our work makes use of the timestamp at which two mobility patterns share the same cell. A case study is also described to give a comparison of the combination measure with other ones.*

## 1. INTRODUCTION

With the development of mobile computing and wireless communications, discovering knowledge about the movement of various groups of mobile objects in wireless networks has become critical in their mobility prediction. Although different mobile objects express differences in their movement behavior and the nature of their movement but they typically share similarities [1]. Discovering such similarities contributes significantly to predicting the next location of a mobile object based on the behaviors of members in a group.

In our work [2], the prediction of the next location of a mobile object is only based on its own movement history. However, the incompleteness on information of movement history results in the lack of extracted mobility rules and may affect the accuracy of prediction. In this paper, we consider that with knowledge of groups to which a mobile object belongs, one can derive common behaviors among objects during their moving. Therefore, it is possible to predict the next location of an object based on the movement behavior of its group. And in turn, the behavior of a group of mobile objects is determined in terms of the similarity of movement patterns, which represent a homogeneous kind of correlations of behaviors of objects in wireless networks. The analysis of moving objects (i.e., entities whose positions or geometric attributes change over time) has recently attracted a great deal of studies, especially, investigating their trajectories (i.e. paths objects passed through in space and time). Measuring the similarity between trajectories and then clustering them are becoming crucial for movement prediction of mobile objects [3].

Approaches for computing the similarity between sequences in trajectory data of moving objects may be grouped into two classes. On the one hand, the methods based on Euclidean space such as in [4] [5] [6] [7] [8] consider similarity with the Euclidean distance. On the other hand, some works [9] [10] [11] have investigated the spatial and temporal properties of patterns but with interval temporal aspect. And then two patterns passing through the same area at different times are considered to be similar. These measures may be not appropriate to spatiotemporal characteristics of cellular space in wireless network. A more detail description and analysis of these measures are postponed to Section 2.

Our contribution in this paper is to introduce a new similarity measure of mobility patterns. Our approach to constructing a such measure is to deal with two constraints. First, the measure should be based on the characteristics of the mobility in wireless network, in which a location of mobile objects is referenced with the cell identifier. Second, the measure needs to represent spatial and temporal similarities simultaneously. It means that two patterns passing through the same cell at the same time must be considered more similar than the case where there is no common timestamp. Our similarity measure is computed as a weighted combination of spatial and temporal ones. The assertion on its truth will be presented in mathematical proof and then a comparable computation between this measure with some other measures is given in a case study.

The remainder of this paper is constructed as follows. In Section 2, a review of approaches to similarity measurement in trajectory data is introduced. Section 3 presents the model of the mobility in wireless networks and Section 4 investigates a new similarity measure between mobility patterns. In Section 5, we describe a case study for comparing various similarity measures of mobility patterns. Section 6 is some discussions. Finally, Section 7 draws concluding remarks and further work.

## 2. OVERVIEW OF SIMILARITY MEASURES

### 2.1. Similarity Measures for Euclidean Space

The similarity measures for trajectory data in Euclidean space with the referenced location of coordinates have been widely considered in various research areas. Lin et al. [4] proposed a searching method for similar trajectories by focusing on the spatial shapes and comparing spatial shapes of moving object trajectories. Searching algorithms based on evaluating *OWD* (one way distance) in both continuous and discrete cases of the trajectories have been developed. However, their searching scheme *OWD* is time independent and based on Euclidean distance. Therefore, it is not appropriate to comparing two trajectories in cellular space of wireless network.

Another method for measuring the similarity between trajectories based on spatio-temporal representation was introduced by Zeinalipour-Yazti et. al. [5]. They proposed a distributed spatio-temporal similarity measure using the *LCSS* (longest common subsequence) distance. Their approach performs local computations of partial lower and upper bounds at each cell and then combines these partial results to give upper and lower bounds. However, this scheme also assumes Euclidean space, it is difficult to apply it to spatial networks [11]. Some other effective similarity measures were proposed in [6] [7] [8] but they have the same representation of

Euclidean space as [4] [5]. Due to the spatial and temporal properties of wireless networks, these methods are no longer useful.

## 2.2. Similarity Measures without Temporal Aspect

The most important studies of similar trajectories which are suitable for spatial networks are presented in [9] and [10]. These studies consider that geographically close trajectories may not necessarily be similar since the activities implied by nearby landmarks they pass through may be different. Jia-Ching Ying et. al. [9] proposed a novel approach for measuring the semantic similarity between trajectories, namely, *Maximal Semantic Trajectory Pattern Similarity* (MSTP-Similarity). After transforming the geographic trajectory set to the semantic trajectory dataset, they utilized the sequential pattern mining algorithm *Prefix-Span* to mine the frequent semantic trajectories called *semantic trajectory patterns*. And then they used the *Longest Common Sequence* (LCS) of two patterns to represent their longest common part and defined *MSTP-Similarity* based on the *participation ratio* of the common part to a pattern. *MSTP-Similarity* argued that two trajectory patterns are more similar when they have more common parts. However, the disadvantage of this model is that it did not take temporal property into account. For example, two trajectory patterns passing through the same area at different times are considered to be similar. Consequently, it is not suitable to spatio-temporal domains.

A similarity measure based on sequential mobility patterns with time and space has been considered by Pandi et. al. [1]. They stated that the main feature of sequence data is the order of sequential elements. Thus, they proposed a new sequence similarity measure that focused on the ordering feature of sequences. Their evaluation results showed the superiority of the measure compared to other evaluated measures. However, this method has the same problem of temporal property as [9].

## 2.3. Similarity Measures of Networks with Simultaneously Spatiotemporal Consideration

To the best of our knowledge, studying the spatial network constraints by taking both spatial and temporal properties of patterns into account is satisfactory. Firstly, the work given by Tiakas et. al. [12] used the network distance instead of the Euclidean distance. The spatial network is modeled as a directed graph, and the network distance is defined by using algorithms for shortest paths between the nodes of the graph as follows. Let $D_{net}(T_a, T_b)$ be a distance between two trajectories $T_a = (v_{a1}, v_{a2}, ..., v_{am})$ and $T_b = (v_{b1}, v_{b2}, ..., v_{bm})$ of length $m$. Then

$$D_{net}(T_a, T_b) = \frac{1}{m} \sum_{i=1}^{m} d(v_{ai}, v_{bi})$$

$$d(v_{ai}, v_{bi}) = \begin{cases} 0, & \text{if } v_{ai} = v_{bi} \\ \frac{c(v_{ai}, v_{bi}) + c(v_{bi}, v_{ai})}{2 D_G}, & \text{otherwise} \end{cases}$$

where $D_G = \max\{c(v_i, v_j), \forall v_i, v_j \in V(G)\}$ is the diameter of the graph G of the spatial network. In addition to $D_{net}(T_a, T_b)$, they also compute the time similarity $D_{time}(T_a, T_b)$ between two trajectories as follows:

$$D_{time}(T_a, T_b) = \frac{1}{m-1} \sum_{i=1}^{m} \frac{|(T_a[i+1]t - T_a[i]t) - (T_b[i+1]t - T_b[i]t)|}{\max\{(T_a[i+1]t - T_a[i]t), (T_b[i+1]t - T_b[i]t)\}}$$

In order to take both spatial and temporal properties into account, they combined the two distance measures $D_{net}$ and $D_{time}$ into a single one:

$$D_{total}(T_a, T_b) = W_{net}.D_{net}(T_a, T_b) + W_{time}.D_{time}(T_a, T_b)$$

where $W_{net} + W_{time} = 1$

The advantage of this approach is that it utilizes network distance and takes into spatiotemporal aspects account simultaneously. However, geographically close trajectories may not necessarily be similar in some real applications [9]. For example, in Figure 1, although the geographic distance between *Trajectory 1* and *Trajectory 2* is closer than that between *Trajectory 1* and *Trajectory 3*, both *Trajectory 1* and *Trajectory 3* pass through the same locations <School, Park, Restaurant>. Thus, *Trajectory 1* and *Trajectory 3* are considered more similar to each other than to *Trajectory 2*. Such an approach is inconvenient to apply to our model of wireless network.

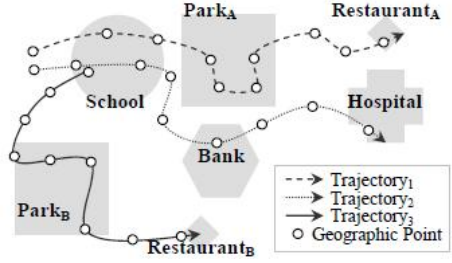

Figure 1. An example of similar trajectories [9]

Secondly, Kang et. al. [13] used cellular space instead of Euclidean space in order to construct the similarity between the patterns of moving objects. According to [13], the location is referenced by cells rather than coordinates in real world. For example, we refer the location of an object in a building by the room number. In order to process a spatial query "Who are in the room 406?", the location referenced by room identifier is more useful than ($x$, $y$) coordinates. Based on this observation, Kang et. al. defined two similarity measures by using LCSS, which had been originally proposed to analyze the similarity between sequences. Furthermore, they simultaneously take spatial and temporal properties of trajectory patterns into account which are not fully considered by LCSS. Firstly, Kang et. al. defined spatial similarity $LCSS_{a,b}(n, m)$ where $a$ and $b$ are two trajectory patterns and $n$ and $m$ are the numbers of cells visited by $a$ and $b$ respectively as follows:

$$LCSS_{a,b}(i,j) = \begin{cases} 0, \text{if } i = 0 \text{ or } j = 0 \\ LCSS_{a,b}(i-1, j-1)+1, \text{if } i > 0, j > 0, a_i.c = b_j.c \\ max(LCSS_{a,b}(i, j-1), LCSS_{a,b}(i-1, j)), \text{otherwise} \end{cases}$$

where $a_i.c$ and $b_j.c$ are the $i$-th and $j$-th visited cell by $a$ and $b$ respectively. The LCSS means the longest common sequence of cells of patterns. A large value of LCSS means that two patterns share a long subsequence of visits at cells and have a high similarity.

In order to overcome the shortcoming of LCSS, [13] proposed a new definition of measure for patterns similarity *Common Visit Time Interval* (CVTI) between two trajectory patterns $a$ and $b$. CVTI implies the sum of time interval that $a$ and $b$ stay at the same cells:

$$CVTI(a,b) = \sum_{i=1, j=1}^{n,m} |a_i.I \cap b_j.I|, \text{ where } a_i.c = b_j.c$$

Thus, their approach focuses on similarity measure with the time interval which is different from the timestamp as in our work.

Finally, one of the most suitable methods for our purpose is defined by Gómez-Alonso et. al. [14]. Their main purpose is to measure a similarity of sequences of events representing the behavior of the user in a particular context such as sequences of web pages visited by a certain user or the personal daily schedule. This measure is based on the comparison of the common elements (i.e. events such as web pages, places, etc.) in two sequences and the positions where they appear. The former computes if the two individuals have done the same things, e.g if they have visited the same web pages or have gone to the same places. The later takes into account the temporal sequence of the events, e.g., if two tourists have visited the place *A* after going to the place *B* or not.

For example, let *T*1 and *T*2 be tourists who have visited some places of the same city: *T*1={*a*, *b*, *c*} and *T*2={*c*, *a*, *b*, *d*}. There are 3 common places and also they have visited *a* before *b*. So they are quite similar. In order to take into account these two issues, the new measure called *Ordering-based Sequence Similarity* (OSS) consists, on one hand, of finding the common elements in the two sequences, and on the other hand, in comparing the positions of the elements in both sequences. Let *i* and *j* be two sequences of items of different lengths, $i = (x_{i,1}, \ldots, x_{i,\text{card}(i)})$ and $j = (x_{j,1}, \ldots, x_{j,\text{card}(j)})$. Let $L = \{l_1, \ldots, l_n\}$ be a set of *n* symbols to represent all the possible elements of those sequences (*L* is called a language). Then, the OSS is defined as:

$$d_{OSS}(i,j) = \frac{f(i,j) + g(i,j)}{card(i) + card(j)}$$

where

$$g(i,j) = card(\{x_{ik} \mid x_{ik} \notin j\}) + card(\{x_{jk} \mid x_{jk} \notin i\})$$

and

$$f(i,j) = \frac{\sum_{k=1}^{n}\left(\sum_{p=1}^{\Delta}\left|i_{(l_k)}(p) - j_{(l_k)}(p)\right|\right)}{\max\{card(i), card(j)\}}$$

where $i_{(l_k)} = \{t \mid i(t) = l_k\}$ and $\Delta = \min\{card(i_{(l_k)}), card(j_{(l_k)})\}$.

The measure OSS has two parts, *g* is counting the number of non common elements, and *f* measures the similarity in the position of the elements in the sequences (the ordering). If two sequences are equal, the result of $d_{OSS}$ is zero, because the positions are always equal (*f* = 0) and there are not uncommon elements (*g* = 0). Conversely, if the two sequences do not share any element, then *g* = *card*(*i*) + *card*(*j*) and *f* = 0, and $d_{OSS}$ is equal to 1 when it is divided by *card*(*i*) + *card*(*j*). The values of $d_{OSS}$ are always between 0 and 1.

However, the ordering between the common elements in the two patterns may be no longer useful for our purpose. For example, let *S*1 and *S*2 be two trajectory patterns: *S*1={(1, $t_1$), (0, $t_3$), (5, $t_4$), (6, $t_6$), (7, $t_9$)} and *S*2={(0, $t_3$), (5, $t_4$), (7, $t_9$)}. Cells 0, 5 and 7 are common in both patterns *S*1 and *S*2. Since these two patterns pass through the same cells at the same times, they must be considered temporal similar, even though they are not similar in OSS measure due to *f*(*S*1, *S*2) = 0.8 > 0. This computing result is obtained as follows: $S1_{(0)} = \{1\}$ and $S2_{(0)} = \{0\}$, so $f_{(0)}(S1, S2) = |1-0| = 1$. Similarly, $f_{(5)}(S1, S2) = |2-1| = 1$ and $f_{(7)}(S1, S2) = |4-2| = 2$.

So $f(S1, S2) = \frac{f_{(0)}(S1,S2) + f_{(5)}(S1,S2) + f_{(7)}(S1,S2)}{\max\{card(S1), card(S2)\}} = \frac{1+1+2}{5} = 0.8$

The problem arisen in the measure given by G´omez-Alonso et. al. [14] is a motivation for considering its extension and that is the purpose of this paper. The next section is devoted to a brief description of our model of mobility patterns [2] [15] that is the basis for our proposed similarity measure.

## 3. MOBILITY MODEL IN WIRELESS NETWORKS

In this paper, it is assumed that the radio coverage region is represented by a hexagonal shaped network (see Figure 2). Each hexagon is a cell which is served by a Base Station (BS) in the communication space. The mobile nodes can travel around the coverage region. The bidirected graph is utilized to illustrate the mobility model of nodes in wireless network. Suppose that $G = (V, E)$ is an unweighted directed graph, where *V* is the set of cells in the coverage region and the set E of edges represents the adjacence between pairs of cells. If two cells, say *A* and *B*, are neighboring cells in the coverage region then *G* has a directed and unweighted edge from *A* to *B* and also from *B* to *A*. These bidirected edges illustrate the fact that a mobile node may move from *A* to *B* or *B* to *A* directly and further may travel around the coverage region corresponding graph *G*. The example network shown in Figure 2 can be modeled by the vertex-set *V* = {0, 1, 2, 3, 4, 5, 6, 7, 8, 9, 10, 11} and the edge-set *E* = {(0, 1), (0, 2), (1, 0), (1, 2), (1, 9), (2, 0), (2, 1), (2, 3), (2, 8), (2, 9), ..., (11, 6), (11, 7), (11, 10)}.

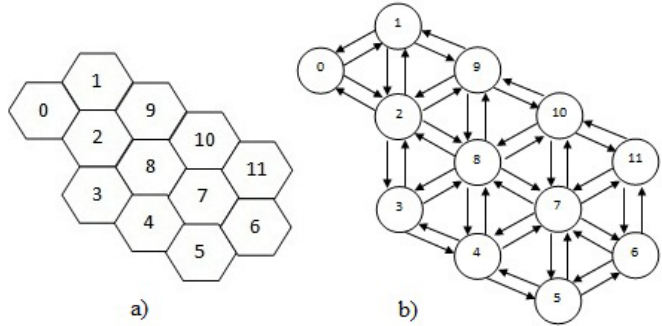

Figure 2. An example coverage region (a) and the corresponding bidirected graph G (b)

## 3.1. Representing Mobility Profiles with Spatiotemporal

Behaviors of mobile users can be characterized in many different ways. In this work, two characteristics which are used to define mobility behaviors are related to location and time-of-day. The following is some discussion of the motivation for using these characteristics.

The location factor indicates that the movement of mobile users often follows a sequence of locations every day. For example, in a campus network, lecturers often move to classrooms, laboratories, library, etc whereas departmental staffs often travel around the administrative offices. Therefore, it is possible to predict the next location of a mobile user based on his own location history.

Table 1. Predefined timestamps

| Timestamps | Time Interval | Timestamps | Time Interval |
|---|---|---|---|
| $t_1$ | 0:00 - 2:14 | $t_7$ | 13:00 - 15:44 |
| $t_2$ | 2:15 - 4:29 | $t_8$ | 15:45 - 17:59 |
| $t_3$ | 4:30 - 6:44 | $t_9$ | 18:00 - 20:14 |
| $t_4$ | 6:45 - 8:59 | $t_{10}$ | 20:15 - 22:29 |
| $t_5$ | 9:00 - 11:14 | $t_{11}$ | 22:30 - 23:59 |
| $t_6$ | 11:15 - 13:29 | | |

The time-of-day factor identifies the importance of the time when a mobile user moves to a location. The mobility behaviours changes as a function of time. In this work, we analyze the individual mobility patterns of lecturers in a campus wireless network. For example, the movement of lecturers depends on the schedule of classes. It means that he will move to the classrooms at the times according to his own teaching schedule. Therefore, it is possible to set the time interval every three teaching periods. A teaching period is 45 minutes so the time interval is 135 minutes. In our work, the predefined timestamps are illustrated in Table 1.

## 3.1. Formalizing Mobility Patterns in Wireless Network

This subsection presents the formalization for modeling mobility patterns. As discussed earlier, the mobile node can travel around the coverage region corresponding to a graph *G*. Let *c* be the ID number of the cell to which the mobile node connected at the timestamp *t*, a point is defined as follows.

**Definition 1.** *Let C and T be two sets of ID cells and timestamps, respectively. The ordered pairs p = (c, t), in which $c \in C$ and $t \in T$, is called a point. Denote P to be the set of all points $P = C \times T = \{(c, t) \mid c \in C \text{ and } t \in T\}$.*

Two point $p_i = (c_i, t_i)$ and $p_j = (c_j, t_j)$ are said to be equivalent if and only if $c_i = c_j$ and $t_i = t_j$. Point $p_i = (c_i, t_i)$ is defined to be earlier than point $p_j = (c_j, t_j)$ if and only if $t_i < t_j$, and it is denoted as $(c_i, t_i) < (c_j, t_j)$ or $p_i < p_j$.

**Definition 2.** *The trajectory of the mobile node is defined as a finite sequence of points $<p_1, p_2, ..., p_k>$ in $C \times T$ space, where $p_j = (c_j, t_j)$ are points for $1 \leq j \leq k$. A sequence composed of k elements is denoted as a k-pattern.*

Note that the value of each timestamp $t_j$ is not unique in a trajectory, i.e. $t_j$ may be equal to $t_i$ iff they are timestamps of two consecutive points of a trajectory. For example $<(c_1, t_1), (c_2, t_2), (c_3, t_2), (c_4, t_4)>$ is a trajectory. The ascending order of points of trajectory is sorted by using *t* as the key.

**Definition 3.** *A mobility pattern $B = <b_1, b_2, ..., b_m>$ is a sub-pattern of another mobility pattern $A = <a_1, a_2, ..., a_n>$, where $a_i$ and $b_j$ are points, written as $B \subset A$, if and only if there exists integers $1 \leq i_1 < ... < i_m \leq n$ such that $b_k = a_{i_k}$, for all k, where $1 \leq k \leq m$. And then, A is called the super-pattern of B.*

For example, given $A = <(c_4, t_2), (c_5, t_3), (c_6, t_4), (c_8, t_5)>$ and $B = <(c_5, t_3), (c_8, t_5)>$. Then B is a sub-pattern of A and conversely, A is super-pattern of B.

## 4. A NEW SIMILARITY MEASURE FOR MOBILITY PATTERNS

Our new similarity measure is motivated by the following requirements. It should be based on the characteristics of wireless networks and may simultaneously reflect similarities of both space and time. Intuitively, two mobility patterns are considered more similar in space if they share more common cells; patterns passing through the same cells at the same times must be considered to be more similar in time than the case they stayed at the different times. In consequence, the similarity measure between two mobility patterns needs to be based on two factors:
- The number of common cells in the two patterns;
- The timestamps of the common cells.

The former allows us to measure the spatial similarity between mobility patterns and the later takes into account the temporal similarity. This section firstly describes similarity measures of mobility patterns in time and space and then makes use of the weighted composition technique for combining two such measures.

**Definition 4.** *Let S be a set of mobility patterns. A similarity measure $D: S \times S \rightarrow [0, 1]$ is a function from a pair of patterns to a real number between zero and one and satisfies the following conditions:*
  (i)    *Reflexivity: for all $P \in S$ $D(P, P) = 0$;*
  (ii)   *Symmetry: for all $P, T \in S$ $D(P, T) = D(T, P)$.*

### 4.1. Spatial Similarity Measure

In this subsection, we make use of the mobility pattern definition and ignoring the time parameter to represent spatial patterns. Suppose that given two mobility patterns $P_a = <c_{a1}, c_{a2}, ..., c_{an}>$ and $P_b = <c_{b1}, c_{b2}, , ..., c_{bm}>$, where $c_{ai} \in V$ and $c_{bj} \in V$, for all *i, j*.

Spatial similarity measure can be defined in terms of spatial dissimilarity between two mobility patterns. The more uncommon cells there are in two patterns, the more spatially dissimilar they are.

**Definition 5.** *Let f: S × S → R be a function representing the number of uncommon cells in two patterns $P_a$ and $P_b$. Then, f is determined by the formula:*

$$f(P_a, P_b) = card(\{c_{ai} \mid c_{ai} \notin P_b\}) + card(\{c_{bi} \mid c_{bi} \notin P_a\})$$

If the two patterns are equal, there are no uncommon cells in these patterns. The result of $f(P_a, P_a)$ is zero. Oppositely, if the two patterns do not share any cells, the result of $f(P_a, P_b)$ is $(n + m)$ where $n$ and $m$ are the length of patterns $P_a$ and $P_b$, respectively.

A spatial similarity measure between two patterns is then defined as follows:

**Definition 6.** *The spatial similarity measure $D_{space}(P_a, P_b)$ between two patterns $P_a$ and $P_b$ with length n and m, respectively, is defined as follows:*

$$D_{space}(P_a, P_b) = \frac{f(P_a, P_b)}{n + m}$$

**Proposition 1.** *The function $D_{space}(P_a, P_b)$ is a similarity measure.*

**Proof.** It is clear that

$$0 \leq card(\{c_{ai} \mid c_{ai} \notin P_b\}) \leq n \text{ and } 0 \leq card(\{c_{bi} \mid c_{bi} \notin P_a\}) \leq m$$

By summation:

$$0 \leq f(P_a, P_b) = card(\{c_{ai} \mid c_{ai} \notin P_b\}) + card(\{c_{bi} \mid c_{bi} \notin P_a\}) \leq (n + m).$$

Thus, $0 \leq D_{space}(P_a, P_b) = \dfrac{f(P_a, P_b)}{n + m} \leq 1$

It is easy to check the reflexivity. When $P_a = P_b$, the result of $D_{space}(P_a, P_b)$ is zero due to $f(P_a, P_b) = 0$. Additionally, if the two patterns do not share any cell, then $f(P_a, P_b) = (n + m)$ and $D_{space}(P_a, P_b)$ is equal to 1 when it is divided by $(n + m)$.

The symmetric property is proven as follows. We have:

$$f(P_a, P_b) = card(\{c_{ai} \mid c_{ai} \notin P_b\}) + card(\{c_{bi} \mid c_{bi} \notin P_a\})$$

and

$$f(P_b, P_a) = card(\{c_{bi} \mid c_{bi} \notin P_a\}) + card(\{c_{ai} \mid c_{ai} \notin P_b\})$$

then $f(P_a, P_b) = f(P_b, P_a)$

In addition, we have also:

$$D_{space}(P_a, P_b) = \frac{f(P_a, P_b)}{n + m}$$

and

$$D_{space}(P_b, P_a) = \frac{f(P_b, P_a)}{m + n}$$

Due to $f(P_a, P_b) = f(P_b, P_a)$, $D_{space}(P_a, P_b)$ is equal to $D_{space}(P_b, P_a)$. Thus, symmetry property has been proven. The proposition is proved.

Determining the spatial similarity between two patterns is presented in **Algorithm 1**. First of all, counting the number of cells in one pattern but not in another (Step 1-2) using **Algorithm 2**, then calculating the length of patterns (Step 3-4). Applying the formulas in Definition 5 and Definition 6 to calculate the values of function $f$ and $D_{space}$ respectively (Step 5-6).

**Algorithm 1** Spatial Similarity

**Input**: two patterns $P_a$ and $P_b$
**Output**: the spatial similarity between $P_a$ and $P_b$, $D_{space}(P_a, P_b)$

```
1. g₁ ← card(Pa, Pb)
2. g₂ ← card(Pb, Pa)
3. n  ← length(Pa)
4. m  ← length(Pb)
5. f(Pa, Pb)  ← g₁ + g₂
6. Dspace   ← f / (n + m)
7. return Dspace
```

**Algorithm 2** Uncommon Cells Counting Algorithm

**Input**: two patterns $P_a$ and $P_b$
**Output**: number of cells in $P_a$ but not in $P_b$, $card(P_a, P_b)$

```
24. card = 0
25. n ← length(Pa)
26. m ← length(Pb)
27. for all cells c_i ∈ Pa do
28.   j = 1
29.   repeat
30.     c_j ← Pb.c[j]
31.     if c_i = c_j then
32.       break
33.     else
34.       j = j + 1
35.     end if
36.   until j = m
37.   if j = m then
38.     card = card + 1
39.   end if
40. end for
41. return card
```

## 4.2. Temporal Similarity Measure

In reality, the mobility behaviors of mobile objects typically change as a function of time. It means that the temporal property of mobility patterns identifies the importance of the time when the mobile object moves from a location to the other one [2]. Therefore, it is necessary to define temporal similarity measure by means of the temporal dissimilarity between two mobility patterns. Our approach is based on intuition that two patterns must be considered temporally similar when they pass through the same cells at the same time.

For example, let $S1 = \{(1, t_1), (0, t_3), (5, t_4), (6, t_6), (7, t_9)\}$ and $S2 = \{(0, t_3), (5, t_4), (7, t_9)\}$ be two mobility patterns. In this context, there are 3 common cells 0, 5 and 7, and they have passed through cell 0 at $t3$, cell 5 at $t4$ and cell 7 at $t9$. Therefore, two these patterns are considered temporally similar. In order to determine the temporal dissimilarity between patterns, we need to calculate the total of temporal difference between the timestamps of the common cells in two patterns. The smaller the total time difference is, the more temporally similar the two patterns are. In the following, we will formalize these statements.

Suppose that $T$ is the set of predefined timestamps. Let $P_a = <(c_{a1}, t_{a1}), (c_{a2}, t_{a2}), …, (c_{an}, t_{an})>$ and $P_b = <(c_{b1}, t_{b1}), (c_{b2}, t_{b2}), …, (c_{bm}, t_{bm})>$ be two mobility patterns, where for all $i, j$ $c_{ai}, c_{bi} \in V$ and $t_{aj}, t_{bj} \in T$. The temporal similarity measure between patterns is defined as follows:

**Definition 7.** *The temporal similarity measure $D_{time}(P_a, P_b)$ between two patterns $P_a$ and $P_b$ with length n and m, respectively, is given by*

$$D_{time}(P_a, P_b) = \frac{1}{k} \sum_{i=1, j=1}^{n,m} \frac{|t_{ai} - t_{bj}|}{\max(t_{ai}, t_{bj})} \text{ where } c_{ai} = c_{bj}$$

*where k is the number of common cells of $P_a$ and $P_b$.*

**Proposition 2.** *The function $D_{time}(P_a, P_b)$ is the similarity measure.*

**Proof.** Clearly, $D_{time} \geq 0$. We will prove that $D_{time} \leq 1$. Since $|t_{ai} - t_{bj}| \leq max(t_{ai}, t_{bj})$ for all $t_{ai}, t_{bj} \in T$, $\frac{|t_{ai} - t_{bj}|}{max(t_{ai}, t_{bj})} \leq 1 \, \forall t_{ai}, t_{bj} \in T$. According to Definition 7, we just compute the time difference between two timestamps $t_{ai}$ and $t_{bj}$ of common cell ($c_{ai} = c_{bj}$). Let $k$ be the number of common cells of $P_a$ and $P_b$, then $\sum_{i=1, j=1}^{n,m} \frac{|t_{ai} - t_{bj}|}{max(t_{ai}, t_{bj})} \leq k$. Therefore,

$$D_{time}(P_a, P_b) = \frac{1}{k} \sum_{i=1, j=1}^{n,m} \frac{|t_{ai} - t_{bj}|}{max(t_{ai}, t_{bj})} \leq 1.$$

Clearly, when two patterns $P_a$ and $P_b$ pass through the same cells at the same timestamps, $D_{time}(P_a, P_b) = 0$, due to $|t_{ai} - t_{bj}| = 0$. In the case that either $P_a$ or $P_b$ passes through all cells at timestamp $t_0$, then $\frac{|t_{ai} - t_{bj}|}{max(t_{ai}, t_{bj})} = 1$ and consequently $D_{time}(P_a, P_b) = 1$.

The Symmetry is proven as follows. We have:

$$\frac{|t_{ai} - t_{bj}|}{max(t_{ai}, t_{bj})} = \frac{|t_{bj} - t_{ai}|}{max(t_{bj}, t_{ai})} \, \forall t_{ai}, t_{bj} \in T$$

In addition, we have also:

$$D_{time}(P_a, P_b) = \frac{1}{k} \sum_{i=1, j=1}^{n,m} \frac{|t_{ai} - t_{bj}|}{max(t_{ai}, t_{bj})} \text{ where } c_{ai} = c_{bj}$$

and

$$D_{time}(P_b, P_a) = \frac{1}{k} \sum_{i=1, j=1}^{n,m} \frac{|t_{bj} - t_{ai}|}{max(t_{bj}, t_{ai})} \text{ where } c_{ai} = c_{bj}$$

then

$$D_{space}(P_a, P_b) = D_{space}(P_b, P_a)$$

Thus, symmetric property has been proven. The proposition is proved.

Determining the temporal similarity between two patterns is presented in **Algorithm 3**. First of all, determining and counting the number of the common cells in both patterns (Step 4-8), then calculating the total of temporal difference between the timestamps of each common cell in two patterns (Step 9-16). Applying the formulas in Definition 7 to calculate the value of $D_{time}$ (Step 22).

### 4.3. Composition Similarity Measure

A similarity measure in cellular space of wireless network may be constructed from the temporal and spatial similarities. It is a convex or weighted combination of two similarity measures on space and time that have been presented in the previous subsections.

**Definition 8.** *Let $W_{space}$ and $W_{time}$ be the weighted values of spatial and temporal similarity measures respectively, such that $W_{space} + W_{time} = 1$. The composition similarity measure is defined as follows*

$$D(P_a, P_b) = W_{space} . D_{space}(P_a, P_b) + W_{time} . D_{time}(P_a, P_b)$$

It is easy to prove the following proposition.

**Proposition 3.** *The function $D(P_a, P_b)$ is the similarity measure.*

## 5. A CASE STUDY

The definition of similarity may depend on the type of resemblances among objects. The different similarity measures may reflect the different aspects of data and of their context. Two

**Algorithm 3** Temporal Similarity
**Input**: two patterns $P_a$ and $P_b$
**Output**: the temporal similarity between $P_a$ and $P_b$, $D_{time}(P_a, P_b)$

```
1. i = 1
2. k = 0
3. total = 0
4. repeat
5.    j = 1
6.    repeat
7.       if P_a.c[i] = P_b.c[j] then
8.          k = k + 1
9.          if P_a.t[i] > P_b.t[j] then
10.            difference = P_a.t[i] - P_b.t[j]
11.            max = P_a.t[i]
12.         else
13.            difference = P_b.t[j] - P_a.t[i]
14.            max = P_b.t[j]
15.         end if
16.         total = total + (difference / max)
17.      end if
18.      j = j + 1
19.   until j = m
20.   i = i + 1
21. until i = n
22. D_time = total / k
23. return D_time
```

patterns can be seen very similar in one measure but rather different in the other measure. In order to demonstrate this observation, we perform a computation of the similarity between two trajectory patterns $S_a = \{(1, t_1), (0, t_3), (2, t_4), (8, t_6), (7, t_9)\}$ and $S_b = \{(0, t_3), (2, t_4), (3, t_5), (8, t_6), (4, t_8)\}$ using different similarity measures. The following analysis will use the directed graph $G$ in Figure 2 and the set of predefined timestamps in Table 1 as an example to explain the idea of different similarity measures.

Firstly, we determine the similarity between two patterns $S_a$ and $S_b$ using the measure proposed in [12]. We have:

$$D_G = \max\{c(v_i, v_j), \text{ for all } v_i, v_j \in V(G)\} = 4.$$

$$D_{net}(S_a, S_b) = \frac{1}{m}\sum_{i=1}^{m} d(v_{ai}, v_{bi})$$

$$D_{net}(S_a, S_b) = \frac{d(1,0) + d(0,2) + d(2,3) + d(8,8) + d(7,4)}{5}$$

$$d(1,0) = \frac{c(1,0) + c(0,1)}{2 \times 4} = \frac{2}{8} = 0.25$$

Similarly, $d(0,2)= 0.25$, $d(2,3)= 0.25$, $d(8,8)= 0$, $d(7,4)= 0.25$.

Thus $D_{net}(S_a, S_b) = \frac{0.25 + 0.25 + 0.25 + 0 + 0.25}{5} = 0.2$

We also have:

$$D_{time}(S_a, S_b) = \frac{1}{m-1} \sum_{i=1}^{m} \frac{|(S_a[i+1]t - S_a[i]t) - (S_b[i+1]t - S_b[i]t)|}{\max\{(S_a[i+1]t - S_a[i]t), (S_b[i+1]t - S_b[i]t)\}}$$

$$= \frac{1}{4} \left\{ \frac{|(t_3 - t_1) - (t_4 - t_3)|}{\max\{(t_3 - t_1), (t_4 - t_3)\}} + \frac{|(t_4 - t_3) - (t_5 - t_4)|}{\max\{(t_4 - t_3), (t_5 - t_4)\}} + \frac{|(t_6 - t_4) - (t_6 - t_5)|}{\max\{(t_6 - t_4), (t_6 - t_5)\}} + \frac{|(t_9 - t_6) - (t_8 - t_6)|}{\max\{(t_9 - t_6), (t_8 - t_6)\}} \right\}$$

$$= \frac{1}{4} \left\{ \frac{1}{2} + \frac{0}{1} + \frac{1}{2} + \frac{1}{3} \right\} = 0.333$$

Taking $W_{net} = 0.5$ and $W_{time} = 0.5$, we have

$$D_{total}(S_a, S_b) = W_{net}.D_{net}(S_a, S_b) + W_{time}.D_{time}(S_a, S_b)$$

$$D_{total}(S_a, S_b) = 0.5 \times 0.2 + 0.5 \times 0.333 = 0.267$$

Secondly, we use the method OSS in [14] to measure the similarity between the two patterns. We have:

$$g(S_a, S_b) = card(\{c_{ai} \mid c_{ai} \notin S_b\}) + card(\{c_{bj} \mid c_{bj} \notin S_a\})$$

$$g(S_a, S_b) = 2 + 2 = 4$$

Notice that $S_a$ and $S_b$ have 3 common cells 0, 2 and 8
$S_{a(0)} = \{1\}$ and $S_{b(0)} = \{0\}$, so $f_{(0)}(S_a, S_b) = |1-0| = 1$
Similarly, $f_{(2)}(S_a, S_b) = |2-1| = 1$ and $f_{(8)}(S_a, S_b) = |3-3| = 0$.
So $f(S_a, S_b) = \frac{f_{(0)}(S_a, S_b) + f_{(2)}(S_a, S_b) + f_{(8)}(S_a, S_b)}{\max\{card(S_a), card(S_b)\}} = \frac{1+1+0}{5} = 0.4$

$$d_{OSS}(S_a, S_b) = \frac{f(S_a, S_b) + g(S_a, S_b)}{card(S_a) + card(S_b)} = \frac{0.4 + 4}{10} = 0.44$$

Finally, we compute the similarity between the two patterns based on our measure. We have

$$f(S_a, S_b) = 2 + 2 = 4$$

Then, $D_{space}(S_a, S_b) = \frac{f(S_a, S_b)}{n + m} = \frac{4}{5 + 5} = 0.4$

$$D_{time}(S_a, S_b) = \frac{1}{k} \sum_{i=1, j=1}^{n,m} \frac{|t_{ai} - t_{bj}|}{\max(t_{ai}, t_{bj})} \text{ where } c_{ai} = c_{bj}$$

$$= \frac{1}{3} \left\{ \frac{|t_3 - t_3|}{\max(t_3, t_3)} + \frac{|t_4 - t_4|}{\max(t_4, t_4)} + \frac{|t_6 - t_6|}{\max(t_6, t_6)} \right\} = 0$$

Taking $W_{space} = 0.5$ and $W_{time} = 0.5$, we get the result

$$D_{total}(S_a, S_b) = W_{space}.D_{space}(S_a, S_b) + W_{time}.D_{time}(S_a, S_b)$$

$$D_{total}(S_a, S_b) = 0.5 \times 0.4 + 0.5 \times 0 = 0.2$$

## 6. DISCUSSIONS

As presented in Section 2, there are various approaches in computing the similarity between sequences in trajectory data of moving objects. Most of them have focused on Euclidean space or not fully exploited the spatial and temporal properties of mobility patterns. For that reason, they are not appropriate for measuring similarity of spatio-temporal patterns in wireless network. This section is devoted to an additional comparable discussion between our similarity measure and some closely related approaches.

First, the work in [12] used the network distance instead of the Euclidean distance. This method considered simultaneously spatial and temporal aspects. However, geographically close trajectories may not necessarily be similar since the activities implied by nearby landmarks they pass through may be different. In wireless networks, the more common cells two trajectory patterns share, the more similar they must be considered.

Second, Kang et. al. [13] proposed measure $LCSS_{a,b}(n, m)$ based on cellular space instead of Euclidean space. Since $LCSS_{a,b}(n, m)$ is given in recursive form, it may result in large computational complexity. Moreover, the similarity measure CVTI [13] is concerned with time interval, our one is based on the model of timestamp in wireless network.

Finally, the method that is most closely related to our approach has been given by C. Gomez-Alonso et al. [14]. Their approach computes the spatial similarity measure for sequences of events based on the number of common elements in these sequences. However, the temporal aspect in their approach is the ordering between the common elements in two sequences, rather than the timestamp as in our model.

In general, our proposed similarity measure differs from the existing measures in two aspects. First, due to the properties of wireless networks, the spatial similarity measure is based on the number of common elements in two patterns. It means that two trajectory patterns must be considered more similar when they share more common cells. Second, our temporal similarity measure takes into account the context whether two patterns passing through the same cells at the same times or not. Intuitively, two patterns passing through the same cells at the same times must be considered more temporal similar than the case they stayed at the different times. Our research results on spatiotemporal similarity measure in this work will be utilized to develop an improvement of k-means algorithm [16] [17] [18] for clustering mobility patterns.

## 7. CONCLUSIONS AND FUTURE WORK

In this paper, we have proposed a similarity measure for spatiotemporal mobility patterns of wireless network. We have also presented the model of user's mobility patterns in wireless network developed by ourselves based on which the similarity measure has been constructed. Our measure was concerned with characteristics on both time and space of mobility patterns represented in this model. We have mathematically defined two temporal and spatial similarity measures and then made use of a weighted combination to integrate these partial similarity measures. The truth of this formulation has been proved in mathematics. We have also described a case study for comparing our proposed measure with the other ones. In our future work, we are going to develop a clustering algorithm of mobility patterns based on this similarity measure and to utilize this algorithm to construct a system for supporting prediction of mobile users.

## REFERENCES

[1] Somayeh, Dodge, (2011) "Exploring Movement Similarity Analysis of Moving Objects", *Doctor thesis*, University of Zurich, Switzerland.

[2] Thuy Van, T. Duong & Dinh Que, Tran, (2012) "An Effective Approach for Mobility Prediction in Wireless Network based on Temporal Weighted Mobility Rule", *International Journal of Computer Science and Telecommunications*, Vol. 3, Issue 2.

[3] Hwang, J., Kang, H. & Li, K., (2005) "Spatio-Temporal Similarity Analysis between Trajectories on Road Networks", *Second International Workshop on Conceptual Modeling for Geographic Information Systems* (CoMoGIS), Springer-Verlag, pp. 280-289, Austria.

[4] Lin, B. & Su, J. (2005) "Shapes Based Trajectory Queries for Moving Objects", *Proc. of the 13th annual ACM international workshop on Geographic information systems* (GIS'05), pp. 21–30.

[5] Zeinalipour-Yazti, D., Song Lin, S. & Gunopulos, D., (2006) "Distributed Spatio-Temporal Similarity Search", *Proc. of the 15th ACM International Conference on Information and Knowledge Management* (CIKM), pp. 14-23, USA.

[6] Vlachos, M., Kollios, G. & Gunopulos, D., (2002) "Discovering Similar Multidimensional Trajectories", *Proc. of the 18th ICDE, IEEE Computer Society Press, Los Alamitos*, pp. 673–684.

[7] Sakurai, Y., Yoshikawa, M. & Faloutsos, C., (2005) "FTW: Fast Similarity Search Under the Time Warping Distance", *Proc. of the 24th ACM SIGACT-SIGMOD-SIGART Symposium on Principles of Database Systems* (PODS), pp. 326–337, USA.

[8] Chen, L., Ozsu, M. T. & Oria, V., (2005) "Robust and Fast Similarity Search for Moving Object Trajectories", *Proc. of the ACM SIGMOD International Conference on Management of Data* (ACM SIGMOD), pp. 491–502, USA.

[9] Ying, J. Jia Ching, Lu, E. Hsueh-Chan & Lee, Wang-Chien, (2010) "Mining user similarity from semantic trajectories", *Proc. of the ACM International Workshop on Location Based Social Networks*, pp. 19-26, USA.

[10] Mohammad H. Pandi, Omid Kashefi & BehrouzMinaei, (2011) "A Novel Similarity Measure for Sequence Data", *Journal of Information Processing Systems*, Vol.7, No.3.

[11] Chang, Jae-Woo, Bista, Rabindra, Kim, Young-Chang & Kim, Yong-Ki, (2007) "Spatio-temporal Similarity Measure Algorithm for Moving Objects on Spatial Networks", *Proc. of the International Conference on Computational Science and Its Applications* (ICCSA), Springer-Verlag Berlin Heidelberg, pp. 1165-1178, Malaysia.

[12] Tiakas, E., Papadopoulos, A. N., Nanopoulos, A., Manolopoulos, Y., Stojanovic, D. & Djordjevic-Kajan, A., (2009) "Searching for similar trajectories in spatial networks", *the Journal of Systems and Software, Vol. 82, No. 5*, pp. 772-788.

[13] Kang, HyeYoung, Kim, JoonSeok & Li, KiJoune, (2009) "Similarity measures for trajectory of moving objects in cellular space", *Proc. of the 2009 ACM Symposium on Applied Computing* (SAC), pp. 1325-1330, USA.

[14] G´omez-Alonso, Cristina & Valls, Aida, (2008) "A Similarity Measure for Sequences of Categorical Data Based on the Ordering of Common Elements", *Proc. of the 5th International Conference on Modeling Decisions for Artificial Intelligence* (MDAI), Springer-Verlag, pp. 134-145, Spain.

[15] Thuy Van T. Duong & Dinh Que Tran, (2011) "Modeling Mobility in Wireless Netwok with Spatiotemporal State", *International Conference in Mathematics and Applications* (ICMA), pp. 147-155, Thailand.

[16] R. U. Payli, K. Erciyes, & O. Dagdeviren, (2011), "Cluster-based Load Balancing Algorithms for Grids", *International Journal of Computer Networks & Communications*, AIRCCSE, Vol. 3, No. 5.

[17] S.Ayyasamy & S.N. Sivanandam, (2010) "A Cluster Based Replication Architecture for Load Balancing in Peer-to-Peer Content Distribution", *International Journal of Computer Networks & Communications*, AIRCCSE, Vol. 2, No. 5.

[18] Raied Salman, Vojislav Kecman, Qi Li, Robert Strack & Erik Test (2011) "Fast k-means Algorithm Clustering", *International Journal of Computer Networks & Communications*, AIRCCSE, Vol.3, No.4.

**Authors**

**Thuy Van T. Duong** received the B.E. degree in Information Technology from Post & Telecommunication Institute of Technology, Viet Nam, in 2004 and the M.Sc. degree in Computer Science from the HOCHIMINH University of Technology, VietNam, in 2008. She is, currently, a Ph.D. candidate at Post & Telecommunication Institute of Technology, Viet Nam. Her research interests include artificial intelligence, data mining, mobile data management and Communication Technology. She is, currently, a lecturer of Information Technology Department of TONDUCTHANG University in HOCHIMINH city, Vietnam.

**Dinh Que Tran** received the M.Sc. Degree in Software Engineering from Melbourne University, Australia, in 1998 and Ph.D in Computer Science from Information Technology Institute, Vietnam, in 2000. In 2001, he works as a research fellowship at Computer and Software Engineering, Calgary University, Canada. His research interests include artificial intelligence, distributed and intelligent computing, data mining, web mining, semantic web, semantic web service and wireless network. He is an author/co-author of many papers published in National/International Journals and Conferences. He is, currently, Associate Professor of Information Technology Faculty, Post & Telecommunication Institute of Technology, Hanoi, Vietnam.

**Cong Hung Tran** received the B.E in electronic and Telecommunication engineering with first class honors from HOCHIMINH University of Technology, VietNam, in 1987. He received M.Eng. Degree and Ph.D in Telecommunications Engineering from HaNoi University of Technology, VietNam, 1998 and 2004, respectively. His research interests include B – ISDN performance parameters and measuring methods, QoS in high speed networks, MPLS. He is, currently, Associate Professor of Faculty of Information Technology II, Posts and Telecoms Institute of Technology in HOCHIMINH, VietNam.